# The *Deception Archive as* a repository of standardized datasets for large-scale verbal deception detection research

**Caterina Borgese[1], Riccardo Loconte[2], Lucca Pfründer[2], Bennett Kleinberg[2,3]**

[1] Department of Health Sciences, 'Magna Graecia' University of Catanzaro, Catanzaro, Italy
[2] Department of Methodology and Statistics, Tilburg University, Tilburg, Netherlands
[3]Department of Security and Crime Science, University College London, London, UK

**Abstract:** Computational advances in natural language processing and machine learning have substantially expanded the use of automated approaches in verbal deception research. However, existing verbal deception datasets remain fragmented across disciplines, as they are often difficult to identify, access, and reuse, limiting cross-study comparisons and the development of robust computational models. We present the *Deception Archive*, a curated and standardized repository of publicly available verbal deception datasets built upon a large-scale systematic review. Eligible datasets were retrieved, manually inspected, and standardized following a structured data curation workflow. The current repository comprises 41 publicly available datasets containing 387,995 truthful or deceptive statements. The archive represents a heterogeneous repository (e.g., in data size, topic of investigation, ground-truth, etc.) and a living resource for integrating newly available deception datasets over time. By providing centralized access to standardized datasets and accompanying metadata, the *Deception Archive* facilitates dataset discovery, cross-dataset validation, and reproducible evaluation of computational models.

Corresponding author: Riccardo Loconte, r.loconte@tilburguniversity.edu

## BACKGROUND & SUMMARY

The ability to distinguish truthful from deceptive communication has long been a central challenge across psychology, communication studies, and, more recently, computational sciences. Deception occurs in both everyday and high-stakes contexts, including online marketplaces, employment interviews, and forensic and legal settings. In many cases, inaccurate credibility assessment can have dire legal, financial, and societal consequences, including misleading criminal investigations and wrongful convictions, financial fraud, and the spread of misinformation[1]. Consequently, understanding deceptive communication has become a longstanding research objective across disciplines.

Despite its relevance, decades of research have shown that accurately distinguishing truthful from deceptive statements remains a challenging task for human judges, who typically perform only slightly above the chance level when they base their judgments on intuition[2,3]. In the search for more substantiated approaches to understand and detect deception, theoretical frameworks grounded in verbal cues have emerged as the most promising because deception is expected to affect how information is generated, organized, and communicated[4–7]. For example, the Cognitive Load Theory posits that fabricating information imposes greater cognitive demands than telling the truth because deceivers must suppress the truth, construct an alternative account, and maintain internal consistency[8]. These additional cognitive demands are expected to influence the structure, complexity, and amount of detail contained in verbal reports[1,9]. Reality Monitoring theory proposes that truthful accounts, being derived from genuine memories, contain richer perceptual, contextual, and sensory information than fabricated recollections of events[10]. Similarly, the Verifiability Approach suggests that deceivers strategically regulate the information they disclose by avoiding details that could be independently verified while maintaining overall plausibility[11,12]. Although these frameworks differ in their underlying assumptions, they converge on the idea that deception affects verbal communication and leaves traces in language in the form of verbal cues that can, in turn, be used to predict deception.

This reliance on verbal cues has fuelled an interest among deception research in natural language processing and machine learning research. Rather than relying on manually coded linguistic cues or subjective human judgments, automated approaches have enabled the analysis of large collections of truthful and deceptive statements by extracting linguistic features and have led to predictive modelling approaches adapted from supervised machine learning. Importantly, recent evidence suggests that automated approaches can outperform human judges in deception detection. Together, the availability of linguistic features and the scalability of automated approaches made verbal communication particularly well-suited for computational analysis, leading to a rapid increase in automated verbal deception detection research in the past 25 years[13–15] .

As the field of automated verbal deception detection has grown and computational approaches have become increasingly data-intensive, the primary challenge has gradually shifted from developing analytical techniques to ensuring the availability of suitable datasets. Computational models require large, diverse, and well-documented corpora for training, benchmarking, and external evaluation[13]. Consequently, the quality of the deception detection models depends not only on the learning algorithms employed but also on the properties of the datasets on which they were trained and evaluated[13]. More broadly, a growing emphasis on open science and research reproducibility has further highlighted the importance of transparent, well-documented, and reusable research datasets, particularly in data-intensive research.

Despite the growing number of datasets on verbal deception published over the past two decades[15,16], identifying and accessing these resources remains difficult. Existing datasets differ in characteristics (e.g., language, data collection procedures, size), are dispersed across scientific areas, and are available in non-standardized formats. As a result, researchers seeking to identify appropriate datasets, reproduce previous work, or develop new computational models invest considerable effort in locating, retrieving, and understanding existing resources. Such fragmentation limits reproducibility and, ultimately, the scientific progress within verbal deception research.

To address these challenges, we present the *Deception Archive*, a curated repository of datasets on verbal deception compiled from a large-scale systematic review of the computational verbal deception literature[15]. In the present article, we describe its development, including the structured data curation workflow, and further provide a descriptive overview of the key properties of the included datasets. The *Deception Archive* is a transparent and standardized research resource to support future computational, psychological, and linguistic investigations of verbal deception. The *Deception Archive* is publicly available in an interactive, searchable interface via https://deception-archive.net; all data underlying the repository are available through Open Science Framework (OSF; link: https://osf.io/x5wpr/overview?view_only=b1a69771e2cf4150be10bebf8b3c3391) and Zenodo (DOI: *removed for peer review*).

## METHODS

### Construction of the *Deception Archive*

The construction of the *Deception Archive* followed a systematic workflow consisting of four phases: (1) selection of candidate datasets, (2) dataset retrieval, (3) dataset curation and standardization, and the final (4) repository construction. Figure 1 provides an overview of the complete data curation protocol.

### Phase 1: Selection of Candidate Datasets

The identification of candidate datasets was based on the corpus of publications included in a recent systematic review and meta-analysis[15], which systematically identified and synthesized studies applying computational methods to verbal deception detection. The review was conducted according to PRISMA guidelines and screened an initial 29,023 publications from both peer-reviewed literature and preprints. Screening and eligibility assessment resulted in a final set of 289 eligible records. For those, a comprehensive set of record-level variables was extracted, including methodological characteristics, computational approaches, and dataset-related information (Table 1). Out of all 289 initial records, 10 records were excluded because it was unclear whether the datasets had been newly collected by the authors or reused from previous studies, 20 records were excluded because no information on the ground-truth labels was available, and 13 records were excluded because they were not documented in an academic publication, leaving 246 final eligible records. From this pool of records, we then identified 134 unique datasets for retrieval. Among these, 56 were unique datasets reused across multiple publications, and 78 represented newly developed datasets.

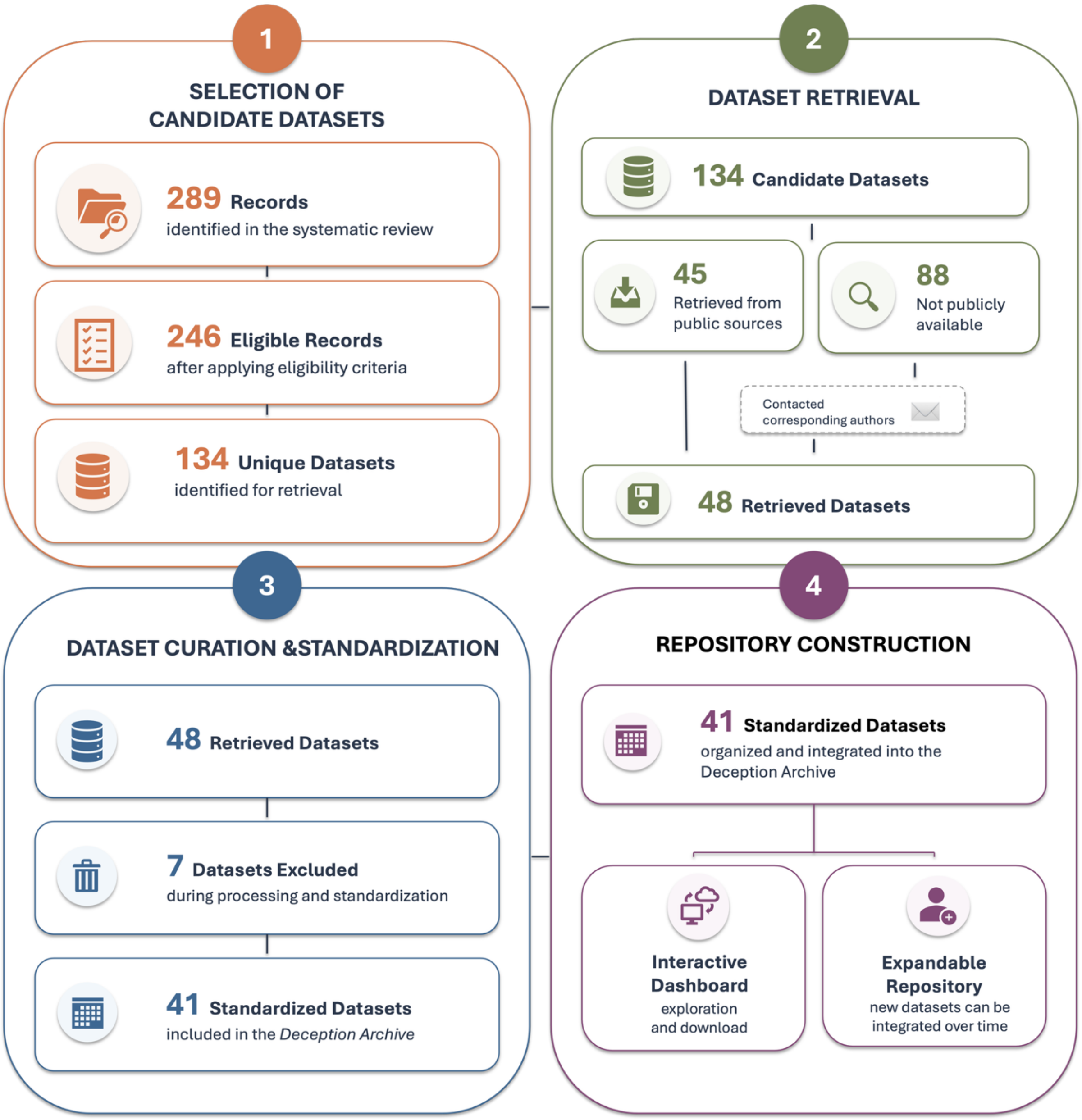

**Figure 1.** Overview of the *Deception Archive* development workflow

### Phase 2: Dataset Retrieval

For each of the 134 datasets retained, a systematic dataset retrieval procedure was used to locate the corresponding datasets. All related publications were first manually inspected for information regarding dataset availability, including direct download links, supplementary materials, or references to external repositories. If no information enabling dataset retrieval was available within the publication, additional searches were conducted across publicly available repositories (such as Open Science Framework, GitHub, or Kaggle), and project websites. A total of 46 datasets were retrieved using this procedure.

For the remaining 88 datasets (i.e., from papers published by 82 corresponding authors), for which no downloadable dataset could be identified through either the publication or additional searches, we contacted the corresponding authors by email to request access to the original data. If the contacted authors did not respond within 14 days, a follow-up reminder was sent two weeks after the initial request. Overall, 11 authors (13.4%) responded to the data request, of whom nine replied after the initial email and two after the reminder. Among those who responded, three provided access to their datasets, four indicated that the original data were no longer available, three reported that the datasets could not be shared because of data-sharing restrictions, and one provided an alternative download link that ultimately proved inaccessible. Combining the 46 datasets retrieved from publicly available sources with the three datasets obtained after contacting the authors resulted in a final collection of 49 datasets, which formed the basis for the subsequent dataset processing and standardization stage.

**Phase 3: Dataset Processing and Standardization**

Each retrieved dataset was manually inspected and processed to harmonize the substantial heterogeneity in file formats, data structures, and variable names. During this process, the accompanying documentation and, where necessary, the corresponding publication were consulted to correctly interpret the dataset structure, variable definitions, and annotation scheme. Each retrieved dataset was first transformed into a standardized format consisting of six core variables: (i) a unique dataset identifier to ensure traceability across the repository; (ii) a participant identifier to preserve information on whether multiple statements originated from the same individual within a dataset; (iii) the original textual content and (iv) veracity label provided by the dataset creators without modification; (v) a standardized binary veracity label (i.e., truthful vs. deceptive) to facilitate cross-dataset analyses, and (vi) the language of the statement text content. Additional dataset-level variables describing dataset characteristics were also reported, including study design (i.e., within- or between-subjects), level of ground-truth (following previous definitions[15]), truthful/deceptive proportion, format of statements (i.e., typed vs. transcribed), macro topic (e.g., Spam) and sub-topic (e.g., email phishing), type of deception (e.g., fabrication, embedded lies, mixed, etc.), and source of the data (e.g., online experiment). The metadata fields extracted and curated for each dataset are summarized in Table 1.

During the standardization process, some retrieved datasets were found to lack the information required to be harmonized within the common data structure of the *Deception Archive*. When missing or ambiguous information was identified, the corresponding authors were contacted to resolve the issue. Datasets were excluded only when these efforts failed to provide the information necessary for reliable standardization. Specifically, datasets were excluded when the original textual data were unavailable ($n$=2 datasets), the ground-truth labels were missing or could not be interpreted unambiguously ($n$=2), and the accompanying documentation was insufficient to reliably reconstruct the original dataset structure ($n$=1), or the dataset represented a duplicate of included datasets (*n=1*). Furthermore, two additional large-scale datasets, namely the Amazon Review Dataset[17] and the Yelp Dataset[18], were not retrieved because their exceptionally large size made them unsuitable for inclusion within the standardized curation workflow adopted for the *Deception Archive.* Applying these criteria resulted in the exclusion of eight datasets. Consequently, the final version of the *Deception Archive* comprised 41 standardized datasets[19–59] harmonized within a common data structure and ready-to-use.

**Phase 4: Repository construction**

The standardized datasets were subsequently integrated into the Deception Archive and organized according to a common data structure, with each dataset accompanied by curated metadata describing its main characteristics. An interactive dashboard was implemented to allow researchers to explore the available datasets, inspect their characteristics, and access the corresponding data for download and subsequent computational analyses. The repository was also designed to accommodate the integration of newly available datasets over time, allowing the Deception Archive to be continuously expanded as additional resources become available.

**Ethics statement**

The Deception Archive compiles previously collected and published datasets involving human participants; no new data were collected for the present study. Ethical review and participant consent for the original data collections were obtained by the authors of the respective source studies, as documented in the corresponding publications. The datasets included in the Archive were either publicly released by their creators or shared with us for this purpose.

**Table 1.** Metadata fields included in the Deception Archive

| Variable | Description | Example |
|---|---|---|
| **Core variables** | | |
| **Participant ID** | Participant identifier preserving within-dataset linkage | HIP_2022_1 |
| **Text** | Original unmodified statement | ''Concerts are my most favorite thing, and my boyfriend knew it. […]'' |
| **Original label** | Original annotation provided by dataset authors | Imagined; recalled |
| **Standardized label** | Binary truthful/deceptive label | Truthful; deception |
| **Language** | Language of the statement | English |
| **Dataset-level variables** | | |
| **Dataset ID** | Unique dataset identifier | HIP_2022 |
| **Publication Year** | Year in which the dataset was published/reported | 2022 |
| Within or between design | Whether data were collected within- or between-subjects. | Between/ Within subject |
| **Ground truth** | Reliability level of the deception labels assigned to the dataset | Clear but not verifiable |
| Documented in academic outlet | Whether the dataset is described in a peer-reviewed publication or other academic source | Yes |
| Number of statements | Total number of textual records included in the dataset | 5535 |
| Truthful / deceptive proportion | Distribution of truthful and deceptive statements | 1.0 |
| Format | Format of the verbal material | Typed |
| Macro topic | Main topic or application domain represented in the dataset. | Past events |
| Sub-topic | Specific topic represented within the macro topic | Past memorable experiences |
| Type of deception | Main deception type represented in the dataset. | Fabrication |
| Source | Experimental or naturalistic origin and study design | Online experiment |

**Data Records**

The data underlying the *Deception Archive* are hosted on Zenodo (DOI: *removed for peer review*) and are also available through the OSF (link: https://osf.io/x5wpr/overview?view_only=b1a69771e2cf4150be10bebf8b3c3391). The deposit contains: (i) 41 standardized dataset files in CSV format (UTF-8 encoded), one per dataset, named by the dataset identifier listed in the metadata (e.g., [BLC_2019.csv]); (ii) a dataset-level metadata table ([metadata.csv]) containing the fields described in Table 1; and (iii) a README file documenting the folder structure and variable definitions. Each standardized dataset file contains the variables described in Table 1. The standardized veracity label is coded [0 = truthful, 1 = deceptive]. The Archive is released under a CC BY 4.0 license.

**Data Overview**

Across these datasets, the Archive contains a total of 387,995 statements, with substantial variability in dataset size, ranging from 62 to 108,688 statements (M = 9,463.29; Median = 1,492.00; SD = 25,704.07). This wide range reflects the considerable heterogeneity of existing deception corpora, encompassing both relatively small experimental datasets and large-scale data collections in online settings.

The distribution of dataset characteristics is summarized in Supplementary Table S1. Most datasets are in English (n = 29, 70.7% of datasets, accounting for 95.5% of all statements), contain typed rather than transcribed text (n = 31, 75.6% of datasets), and originate from experimental or quasi-experimental designs, with fabrication as the predominant deception type (n = 30, 73.2% of datasets) and "clear but non-verifiable" as the most common ground-truth level (n = 22, 53.7% of datasets; see Supplementary Table S2 for definitions). At the same time, the Archive spans nine languages beyond English and a broad range of application domains, from consumer reviews and personal narratives to forensic interviews and digital scams.

**Technical Validation**

The quality of the Deception Archive was assessed throughout the curation workflow described in the Methods. Each retrieved dataset was manually inspected, and the mapping from the original annotations to the standardized binary labels was verified against the accompanying documentation and, where necessary, the original publication. Ambiguities were resolved by contacting the corresponding authors, and datasets that could not be reliably standardized were excluded.

**USAGE NOTES**

The Deception Archive reflects the availability and accessibility of verbal deception datasets at the time of compilation: many datasets described in the literature were no longer retrievable, and datasets lacking the information required for reliable standardization were excluded (see Methods). Researchers who wish to contribute new or previously unavailable datasets to future releases of the Archive are invited to contribute their dataset by reaching out the authors or add their dataset directly on https://deception-archive.net/contribute.

When using the Deception Archive, two limitations should be acknowledged. First, the *Deception Archive* includes only datasets that we could successfully retrieve, interpret, and standardize based on the available materials. Consequently, some datasets described in the literature could not be incorporated because the necessary data or documentation were unavailable. Second, the current version of *the Deception Archive* focuses on datasets originating from the automated verbal deception detection literature. Other valuable verbal deception datasets, including those developed outside this research area (e.g., manually annotated verbal cues of deception), are not currently included but could be incorporated in future releases. Taken together, while the *Deception Archive* represents the largest concentrated repository of verbal deception research datasets based on the hitherto most extensive systematic review in the field, it is far from a comprehensive collection of all datasets used for studying verbal deception. Importantly, the open structure of the Deception Archive allows researchers to contribute additional datasets over time, supporting its continued expansion and development as a shared resource for verbal deception research.

For computational reuse, several properties of the standardized data should be taken into account. The participant identifier allows statements from the same individual to be kept within the same partition when constructing training and test splits, avoiding data leakage in within-subjects datasets. The standardized binary label supports cross-dataset analyses, whereas the original labels preserve finer-grained distinctions where available. Users should further account for the unbalanced truthful/deceptive proportions in some datasets, and for the dominance of English-language material when generalizing across datasets. When individual datasets are used, the original source publications should be cited alongside the Archive.

## Data Availability

The data underlying the *Deception Archive* are publicly available on Zenodo (DOI: *removed for peer review*) and through the Open Science Framework (OSF; https://osf.io/x5wpr/overview?view_only=b1a69771e2cf4150be10bebf8b3c3391) under a CC BY 4.0 license. The deposited files are described in the Data Records section. An interactive, searchable interface is available at https://deception-archive.net.

## Code Availability

No custom code is required to access or use the datasets.

All references indicated with * refers to the datasets included in the *Deception Archive*

# Supplementary Information

**Supplementary Table S1.** Distribution of dataset-level characteristics included in the Deception Archive

| Variable | *Value* | *n (%)* |
|---|---|---|
| ***Language*** | English | 29 (70.7%) |
| | Italian | 3 (7.3%) |
| | Dutch | 1(2.4%) |
| | German | 1(2.4%) |
| | Chinese | 1(2.4%) |
| | Polish | 1(2.4%) |
| | Roman Urdu | 1(2.4%) |
| | Arabic | 1(2.4%) |
| | Bilingual | 2(4.9%) |
| | Multilingual | 1(2.4%) |
| ***Format*** | Typed | 31(75.6%) |
| | Transcribed | 9 (22.0%) |
| | Typed / Transcribed | 1 (2.4%) |
| ***Collection methodology*** | Online experimental | 14(34.1%) |
| | Online naturalistic | 7 (17.1%) |
| | Offline naturalistic | 7 (17.1%) |
| | Online quasi-experimental | 5 (12.2%) |
| | Offline experimental | 5 (12.2%) |
| | Mixed | 3 (7.3%) |
| ***Ground truth*** | Clear but non-verifiable | 22(53.7%) |
| | Directly inferred | 8 (19.5%) |
| | Clear and verifiable | 7 (17.1%) |
| | Indirectly inferred | 3 (7.3%) |
| | Mixed | 1 (2.4%) |
| ***Type of deception*** | Fabrication | 30(73.2%) |
| | Mixed deception types | 9 (22.0%) |
| | Embedded lies | 1 (2.4%) |
| | Exaggeration / minimization | 1 (2.4%) |

**Supplementary Table S2.** Ground-truth classification and definitions

| Level | Description |
|---|---|
| **Clear and verifiable** | Ground truth is fully transparent and can be independently verified for each statement. |
| **Clear but not verifiable** | Ground truth is experimentally assigned but individual statements cannot be externally verified. |
| **Directly inferred** | Ground truth is inferred from strong external indicators. |
| **Indirectly inferred** | Ground truth is based on weaker proxy indicators or subjective judgments. |
| **Mixed** | Dataset combines statements with different ground-truth reliability levels. |

*Note.* Ground truth (GT) refers to the degree of confidence that the truthfulness of statements is established through empirical evidence rather than inference. To characterize the reliability of deception labels across datasets included in the *Deception Archive*, each dataset was assigned to one of the five ground-truth levels described below.